\documentclass[10pt,twocolumn,letterpaper]{article}

\usepackage{cvpr}
\usepackage{times}
\usepackage{epsfig}
\usepackage{graphicx}
\usepackage{amsmath}
\usepackage{amssymb}

\usepackage{algorithm}
\usepackage[noend]{algpseudocode}
\usepackage{tabularx}
\usepackage{booktabs}
\usepackage{xcolor}
\usepackage{dsfont}

\usepackage[pagebackref=true,breaklinks=true,letterpaper=true,colorlinks,bookmarks=false]{hyperref}

\cvprfinalcopy % *** Uncomment this line for the final submission

\def\cvprPaperID{1770} % *** Enter the CVPR Paper ID here

\usepackage{color}

\newcommand{\done}{\texttt{Done} }
\newcommand{\lexp}{\mathcal{L}_\text{int}}
\newcommand{\lnav}{\mathcal{L}_\text{nav}}
\newcommand{\SAVN}{SAVN }
\newcommand{\SAVNnospace}{SAVN}
\newcommand{\lgf}{$L \geq 5$}

\begin{document}

%%%%%%%%% TITLE
\title{Learning to Learn How to Learn: \\
Self-Adaptive Visual Navigation using Meta-Learning}

\author{Mitchell Wortsman$^1$, Kiana Ehsani$^2$, Mohammad Rastegari$^1$, Ali Farhadi$^{1,2}$, Roozbeh Mottaghi$^1$\\
$^1$ PRIOR @ Allen Institute for AI, $^2$ University of Washington}

\maketitle
\thispagestyle{empty}

%%%%%%%%% ABSTRACT
\begin{abstract}
Learning is an inherently continuous phenomenon. When humans learn a new task there is no explicit distinction between training and inference. As we learn a task, we keep learning about it while performing the task. What we learn and how we learn it varies during different stages of learning. Learning how to learn and adapt is a key property that enables us to generalize effortlessly to new settings. This is in contrast with conventional settings in machine learning where a trained model is frozen during inference. In this paper we study the problem of learning to learn at both training and test time in the context of visual navigation. A fundamental challenge in navigation is generalization to unseen scenes.
In this paper we propose a self-adaptive visual navigation method (\SAVNnospace) which learns to adapt to new environments without any explicit supervision. Our solution is a meta-reinforcement learning approach where an agent learns a self-supervised interaction loss that encourages effective navigation. Our experiments, performed in the AI2-THOR framework, show major improvements in both success rate and SPL for visual navigation in novel scenes. Our code and data are available at: \url{https://github.com/allenai/savn}.
\end{abstract}

%%%%%%%%% BODY TEXT
\vspace{-0.4cm}
\section{Introduction}
Learning is an inherently continuous phenomenon. We learn further about tasks that we have already learned and can learn to adapt to new environments by interacting in these environments. There is no hard boundary between the training  and the testing phases while we are learning and performing tasks: we learn as we perform. This stands in stark contrast with many modern deep learning techniques, where the network is frozen during inference. 

What we learn and how we learn it varies during different stages of learning. To learn a new task we often rely on explicit external supervision. After learning a task, we further learn as we adapt to new settings. This adaptation does not necessarily need explicit supervision; we often do this via interaction with the environment.  

\begin{figure}[tp]
    \centering
    \includegraphics[width=1\linewidth]{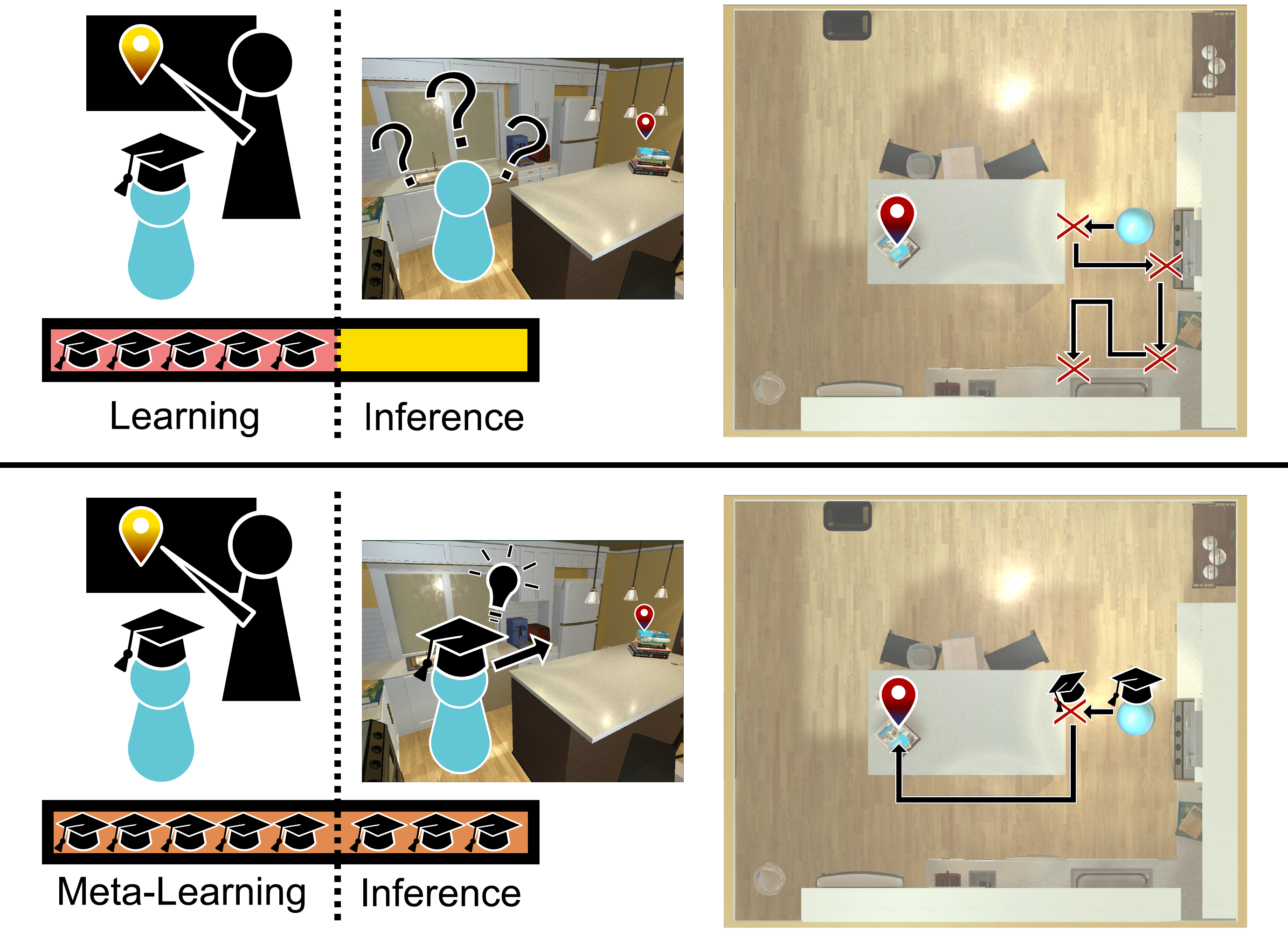}
    \caption{Traditional navigation approaches freeze the model during inference (top row); this may result in difficulties generalizing to unseen environments. In this paper, we propose a meta-reinforcement learning approach for navigation, where the agent learns to adapt in a self-supervised manner (bottom row). In this example, the agent learns to adapt itself when it collides with an object once and acts correctly afterwards. In contrast, a standard solution (top row) makes multiple mistakes of the same kind when performing the task.}
    \label{fig:teaser}
\end{figure}

In this paper, we study the problem of learning to learn and adapt at both training and test time in the context of visual navigation; one of the most crucial skills for any visually intelligent agent. The goal of visual navigation is to move towards certain objects or regions of an environment.  A key challenge in navigation  is generalizing to a scene that has not been observed during training, as the structure of the scene and appearance of objects are unfamiliar. In this paper we propose a self-adaptive visual navigation (\SAVNnospace) model which learns to adapt during inference without any explicit supervision using an interaction loss (Figure \ref{fig:teaser}).  

Formally, our solution is a meta-reinforcement learning approach to visual navigation, where an agent learns to adapt through a self-supervised interaction loss. Our approach is inspired by  gradient based meta-learning algorithms that learn quickly using a small amount of data \cite{finn17}. In our approach, however, we learn quickly using a small amount of self-supervised interaction. In visual navigation, adaptation is possible without access to any reward function or positive example. As the agent trains, it learns a self-supervised loss that encourages effective navigation. During training, we encourage the gradients induced by the self-supervised loss to be similar to those we obtain from the supervised navigation loss. The agent is therefore able to adapt during inference when explicit supervision is not available.

In summary, during both training and testing, the agent modifies its network while performing navigation. This approach differs from traditional reinforcement learning where the network is frozen after training, and contrasts with supervised meta-learning as we learn to adapt to new  environments during inference without access to rewards.

We perform our experiments using the AI2-THOR \cite{ai2thor} framework. The agent aims to navigate to an instance of a given object category (e.g., \emph{microwave}) using only visual observations. We show that \SAVN outperforms the non-adaptive baseline in terms of both success rate (40.8 vs 33.0) and SPL (16.2 vs 14.7).  Moreover, we demonstrate that learning a self-supervised loss provides improvement over hand-crafted self-supervised losses. Additionally, we show that our approach outperforms memory-augmented non-adaptive baselines. 

\vspace{-0.3cm}
\section{Related Work}
\noindent\textbf{Deep Models for Navigation.} Traditional navigation methods typically perform planning on a given map of the environment or build a map as the exploration proceeds \cite{matthies87,thrun98,kidono2002,konolige06,cummins07,blosch10}. Recently, learning-based navigation methods (e.g., \cite{zhu17,gupta17,mirowski17}) have become popular as they \emph{implicitly} perform localization, mapping, exploration and semantic recognition end-to-end. 

Zhu et al. \cite{zhu17} address target-driven navigation given a picture of the target. A joint mapper and planner has been introduced by \cite{gupta17}. \cite{mirowski17} use auxiliary tasks such as loop closure to speed up RL training for navigation. We differ in our approach as we adapt dynamically to a novel scene. \cite{savinov18} propose the use of topological maps for the task of navigation. They explore the test environment for a long period to populate the memory. In our work, we learn to navigate without an exploration phase. \cite{kahn18} propose a self-supervised deep RL model for navigation. However, no semantic information is considered. \cite{mousavian18} learn navigation policies based on object detectors and semantic segmentation modules. We do not rely on heavily supervised detectors and learn from a limited number of examples. \cite{yang18,wu18} incorporate semantic knowledge to better generalize to unseen scenarios. Both of these approaches dynamically update their manually defined knowledge graphs. However, our model learns which parameters should be updated during navigation and how they should be updated. Learning-based navigation has been explored in the context of other applications such as autonomous driving (e.g., \cite{chen15}), map-based city navigation (e.g., \cite{brahmbhatt17}) and game play (e.g., \cite{wu17}). Navigation using language instructions has been explored by various works \cite{anderson18b,chaplot18,hermann17,yu18,misra17}. Our goal is different since we focus on using meta-learning to more effectively navigate new scenes using only the class label for the target.

\noindent\textbf{Meta-learning.} Meta-learning, or learning to learn, has been a topic of continued interest in machine learning research \cite{thrun98b, schmidhuber97}. More recently, various meta-learning techniques have pushed the state of the art in low-shot problems across domains \cite{finn17, mishra17, duan16}.

Finn et al.~\cite{finn17} introduce Model Agnostic Meta-Learning (MAML) which uses SGD updates to adapt quickly to new tasks. This gradient based meta-learning approach may also be interpreted as learning a good parameter initialization such that the network performs well after only a few gradient updates. \cite{li18} and \cite{yu18one-shot} augment the MAML algorithm so that it uses supervision in one domain to adapt to another. Our work differs as we do not use supervision or labeled examples to adapt. %Additionally we do not change domains.

\begin{figure*}[tp]
    \centering
    \includegraphics[width=37pc]{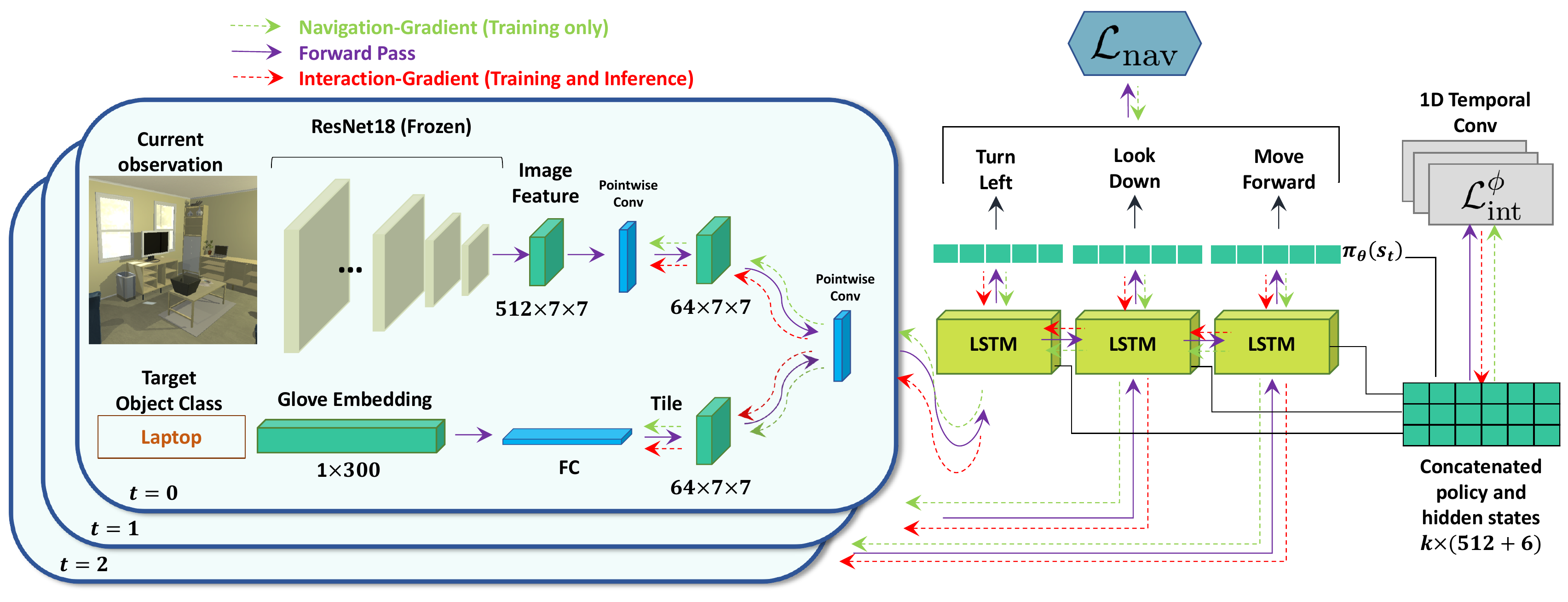}
    \caption{\textbf{Model overview.} Our network optimizes two objective functions, 1) self-supervised interaction loss $\lexp^\phi$ and 2) navigation loss $\lnav$. The inputs to the network at each time $t$ are the egocentric image from the current location and word embedding of the target object class. The network outputs a policy $\pi_\theta(s_t)$. During training, the interaction and navigation-gradients are back-propagated through the network, and the parameters of the self-supervised loss are updated at the end of each episode using navigation-gradients. At test time the parameters of the interaction loss remain fixed while the rest of the network is updated using interaction-gradients. Note that the green color in the figure represents the intermediate and final outputs.}
    \label{fig:model}
    \vspace{-0.4cm}
\end{figure*}

Xu et al. \cite{xu18} use meta-learning to significantly speed up training by encouraging exploration of the state space outside of what the actor's policy dictates. Additionally, \cite{gupta18} use meta-learning to augment the agent's policy with structured noise. At inference time, the agent is able to better adapt from a few episodes due to the variability of these episodes. Our work instead emphasizes self-supervised adaptation while executing a single visual navigation task. Neither of these works consider this domain.

Clavera et al.~\cite{clavera18} consider the problem of learning to adapt to unexpected perturbations using meta-learning. Our approach is similar as we also consider the problem of learning to adapt. However, we consider the problem of visual navigation and adapt via a self-supervised loss. 

Both \cite{houthooft18} and \cite{yu18one-shot} learn an objective function. However, \cite{houthooft18} use evolutionary strategies instead of meta-learning. Our approach for learning a loss is inspired by and similar to \cite{yu18one-shot}. However, we adapt in the same domain without explicit supervision while they adapt across domains using a video demonstration.

\noindent\textbf{Self-supervision.} Different types of self-supervision have been explored in the literature \cite{agrawal15,jayaraman15,doersch15,wang15,zhang16,pinto16,pathak16,owens16}. Some works aim to maximize the prediction error in the representation of future states \cite{pathakICMl17curiosity,stadie15}. In this work, we \emph{learn} a self-supervised objective which encourages effective navigation.

\section{Adaptive Navigation}

In this section, we begin by formally presenting the task and our base model without adaptation.  We then explain how to incorporate adaptation and perform training and testing in this setting.

\subsection{Task Definition} \label{sec:task}

Given a target object class, e.g. \emph{microwave}, our goal is to navigate to an instance of an object from this class using only visual observations.

Formally, we consider a set of scenes $\mathcal{S} = \{S_1,...,S_n\}$ and target object classes $\mathcal{O} = \{o_1,...,o_m\}$. A task $\tau \in \mathcal{T}$ consists of a scene $S$, target object class $o \in \mathcal{O}$, and initial position $p$. We therefore denote each task $\tau$ by the tuple $\tau=(S, o, p)$. We consider disjoint sets of scenes for the training tasks $\mathcal{T}_\text{train}$ and testing tasks $\mathcal{T}_\text{test}$. We refer to the trial of a navigation task as an episode.

The agent is required to navigate using only the egocentric RGB images and the target object class (the target object class is given as a Glove embedding \cite{pennington14}). At each time $t$ the agent takes an action $a$ from the action set $\mathcal{A}$ until the termination action is issued by the agent. We consider an episode to be successful if, within certain number of steps, the agent issues a termination action when an object from the given target class is sufficiently close and visible. If a termination action is issued at any other time, then the episode concludes and the agent has failed.

\noindent\subsection{Learning} 
\label{sec:learning}
Before we discuss our self-adaptive approach we begin with an overview of our base model and discuss deep reinforcement learning for navigation in a traditional sense. 

 We let $s_t$, the egocentric RGB image, denote the agent's state at time $t$.
Given $s_t$ and the target object class, the network (parameterized by $\theta$) returns a distribution over the actions which we denote $\pi_\theta(s_t)$ and a scalar $v_\theta(s_t)$. The distribution $\pi_\theta(s_t)$ is referred to as the agent's \emph{policy} while $v_\theta(s_t)$ is the \emph{value} of the state. Finally, we let $\pi_\theta^{(a)}(s_t)$ denote the probability that the agent chooses action $a$. 

We use a traditional supervised actor-critic navigation loss as in \cite{zhu17, mirowski17} which we denote $\lnav$. By minimizing $\lnav$, we maximize a reward function that penalizes the agent for taking a step while incentivizing the agent to reach the target. The loss is a function of the agent's policies, values, actions, and rewards throughout an episode.

The network architecture is illustrated in Figure \ref{fig:model}. We use a ResNet18 \cite{he16} pretrained on ImageNet \cite{deng09} to extract a feature map for a given image. We then obtain a joint feature-map consisting of both image and target information and perform a pointwise convolution.  The output is then flattened and given as input to a Long Short-Term Memory network (LSTM). For the remainder of this work we refer to the LSTM hidden state and agent's internal state representation interchangeably. After applying an additional linear layer  we obtain the policy and value. In Figure \ref{fig:model} we do not show the ReLU activations we use throughout, or reference the value $v_\theta(s_t)$. 

\subsection{Learning to Learn} 

In visual navigation there is ample opportunity for the agent to learn and adapt by interacting with the environment. For example, the agent may learn how to handle obstacles it is initially unable to circumvent. We therefore propose a method in which the agent learns how to adapt from interaction. The foundation of our method lies in recent works which present gradient based algorithms for learning to learn (meta-learning).

\noindent\textbf{Background on Gradient Based Meta-Learning.} We rely on the meta-learning approach detailed by the MAML algorithm \cite{finn17}. The MAML algorithm optimizes for fast adaptation to new tasks. If the distribution of training and testing tasks are sufficiently similar then a network trained with MAML should quickly adapt to novel test tasks. 

MAML assumes that during training we have access to a large set of tasks $\mathcal{T}_\text{train}$ where each task $\tau \in \mathcal{T}_\text{train}$ has a small meta-training dataset $\mathcal{D}_\tau^\text{tr}$ and meta-validation set
$\mathcal{D}_\tau^\text{val}$. For example, in the problem of $k$-shot image classification, $\tau$ is a set of image classes and $\mathcal{D}_\tau^\text{tr}$ contains $k$ examples of each class. The goal is then to correctly assign one of the class labels to each image
in $\mathcal{D}_\tau^\text{val}$. A testing task $\tau \in \mathcal{T}_\text{test}$ then consists of unseen classes.

The training objective of MAML is given by
\begin{equation} \label{eq:maml}
    \min_{\theta} \sum_{\tau \in \mathcal{T}_\text{train}} \mathcal{L}\left(
    \theta - \alpha \nabla_\theta
    \mathcal{L}\left(\theta, \mathcal{D}_\tau^\text{tr} \right),
    \mathcal{D}_\tau^\text{val}
    \right),
\end{equation}
where the loss $\mathcal{L}$ is written as a function of a dataset and the network parameters $\theta$. Additionally, $\alpha$ is the step size hyper-parameter, and $\nabla$ denotes the differential operator (gradient). The idea is to learn parameters $\theta$ such that they provide a good initialization for fast adaptation to test tasks. 
Formally, Equation~\eqref{eq:maml} optimizes for performance on $\mathcal{D}_\tau^\text{val}$ after \emph{adapting} to the task with a gradient step on $\mathcal{D}_\tau^\text{tr}$. Instead of using the network parameters $\theta$ for inference on $\mathcal{D}_\tau^\text{val}$, we use the \emph{adapted} parameters $\theta - \alpha \nabla_\theta
    \mathcal{L}\left(\theta, \mathcal{D}_\tau^\text{tr} \right)$. In practice, multiple SGD updates may be used
to compute the adapted parameters.

\noindent\textbf{Training Objective for Navigation.} 
Our goal is for an agent to be continually learning as it interacts with an environment. As in MAML, we use SGD updates for this adaptation. These SGD updates modify the agent's policy network as it interacts with a scene, allowing the agent to adapt to the scene. We propose that these updates should occur with respect to $\lexp$, which we call an \emph{interaction loss}.  Minimizing $\lexp$ should assist the agent in completing its navigation task, and it can be learned or hand-crafted. For example, a hand-crafted variation may penalize the agent for visiting the same location twice. In order for the agent to have access to $\lexp$ during inference, we use a self-supervised loss. Our objective is then to learn a good initialization $\theta$, such that the agent will learn to effectively navigate in an environment after a few gradient updates using $\lexp$.

For clarity, we begin by formally presenting our method in a simplified setting in which we allow for a single SGD update with respect to $\lexp$. For a navigation task $\tau$ we let $\mathcal{D}_\tau^{\text{int}}$ denote the actions, observations, and internal state representations (defined in Section~\ref{sec:learning}) for the first $k$ steps of the agent's trajectory. Additionally, let $\mathcal{D}_\tau^{\text{nav}}$ denote this same information for the remainder of the trajectory.
Our training objective is then formally given by
\begin{equation} \label{eq:train-woloss}
    \min_{\theta} \sum_{\tau \in \mathcal{T}_\text{train}} \lnav \left(\theta - \alpha \nabla_\theta \lexp\left(\theta, \mathcal{D}_\tau^\text{int} \right),
    \mathcal{D}_\tau^\text{nav} \right),
\end{equation}
which mirrors the MAML objective from Equation \eqref{eq:maml}. However, we have replaced the small training set $\mathcal{D}_\tau^{\text{tr}}$ from MAML with an interaction phase. The intuition for our objective is as follows: at first we interact with the environment and then we adapt to it. More specifically, the agent interacts with the scene using
the parameters $\theta$. After $k$ steps an SGD update with respect to the self-supervised loss is used to obtain the 
\emph{adapted} parameters $\theta - \alpha \nabla_\theta \lexp\left(\theta, \mathcal{D}_\tau^\text{int} \right)$.

In domain adaptive meta-learning, two separate
losses are used for adaptation from one domain to another \cite{li18, yu18one-shot}. A similar objective to Equation \eqref{eq:train-woloss}
is employed by \cite{yu18one-shot} for one-shot imitation from observing humans.
Our method differs in that we are learning how to adapt in the same domain through self-supervised interaction.

As in $\cite{li18}$, a first order Taylor expansion provides intuition for our training objective. Equation \eqref{eq:train-woloss} is approximated by
\begin{equation} \label{eq:taylor}
\begin{split}
    \min_{\theta} \sum_{\tau \in \mathcal{T}_\text{train}} &\lnav \left(\theta, \mathcal{D}_\tau^\text{nav}  \right) 
    \\
    &- \alpha \left\langle \nabla_\theta \lexp\left(\theta, \mathcal{D}_\tau^\text{int} \right)
    , \nabla_\theta \lnav \left(\theta, \mathcal{D}_\tau^\text{nav}  \right)
    \right\rangle,
\end{split}
\end{equation}
where $\langle \cdot, \cdot \rangle$ denotes an inner product. We are therefore learning to minimize the navigation loss while \emph{maximizing} the \emph{similarity} between
the gradients we obtain from the self-supervised interaction loss and the supervised
navigation loss. 
If the gradients we obtain from both losses are similar, then we are able to continue ``training" during inference when we do not have access to $\lnav$. However, it may be difficult to choose $\lexp$ which allows for similar gradients.
This directly motivates \emph{learning} the self-supervised interaction loss. 

\noindent \subsection{Learning to Learn How to Learn} \label{sec:learnedloss}

We propose to \emph{learn} a self-supervised interaction objective that is explicitly tailored to our task. Our goal is for the agent to improve at navigation by minimizing this self-supervised loss in the current environment. 

During training, we both learn this objective and learn how to learn using this objective. We are therefore ``learning to learn how to learn". As input to this loss we use the agent's previous $k$ internal state representations concatenated with the agent's policy.

Formally, we consider the case
where $\lexp$ is a neural network parameterized by $\phi$, which we denote $\lexp^\phi$.
Our training objective then becomes

\begin{equation} \label{eq:train}
    \min_{\theta, \phi} \sum_{\tau \in \mathcal{T}_\text{train}} \lnav \left(\theta - \alpha \nabla_\theta \lexp^\phi\left(\theta, \mathcal{D}_\tau^\text{int} \right),
    \mathcal{D}_\tau^\text{nav} \right)
\end{equation}
and we freeze the parameters $\phi$ during inference. There is no explicit objective for the learned-loss. Instead, we simply encourage that minimizing this loss allows the agent to navigate effectively. This may occur if the gradients from both losses are similar. In this sense we are training the self-supervised loss to imitate the supervised $\lnav$ loss.

As in \cite{yu18one-shot}, we use one dimensional temporal convolutions for the architecture of our learned loss. We use two layers, the first with $10 \times 1$ filters and the next with $1 \times 1$. As input we concatenate the past $k$ hidden states of the LSTM and the previous $k$ policies. To obtain the scalar objective we take the $\ell_2$ norm of the output. Though we omit the $\ell_2$ norm, we illustrate our interaction loss in Figure \ref{fig:model}.

\begin{algorithm}[tp]
\caption{SAVN-Training$(\mathcal{T}_\text{train}, \alpha, \beta_1, \beta_2, k)$}\label{alg:train}
\begin{algorithmic}[1]
\State{Randomly initialize $\theta, \phi$.}
\While{not converged}
    \For{mini-batch of tasks $\tau_i \in \mathcal{T}_{\text{train}}$}
        \State{$\theta_i \gets \theta$}
        \State{$t \gets 0$}
        \While{termination action is not issued}
            \State{Take action $a$ sampled from $\pi_{\theta_i}(s_t)$}
            %\State{Run for $k$ steps to get self supervised exploration loss $\lexp$.}
            \State{$t \gets t + 1$}
            \If{$t$ is divisible by $k$}
            \State{$\theta_i \gets \theta_i - \alpha \nabla_{\theta_i}
                    \lexp^\phi\left(\theta_i, \mathcal{D}_\tau^{(t,k)}\right)$}
            \EndIf
        \EndWhile
    \EndFor
    \State{$\theta \gets \theta - 
           \beta_1  \sum_i \nabla_\theta \lnav(\theta_i, \mathcal{D}_\tau)$}
    \State{$\phi \gets \phi - 
           \beta_2  \sum_i \nabla_\phi \lnav(\theta_i, \mathcal{D}_\tau)$}
\EndWhile
\State{\textbf{return} $\theta, \phi$}

\end{algorithmic}
\end{algorithm}
\noindent\textbf{Hand Crafted Interaction Objectives.} We also experiment with two variations of simple hand crafted interaction losses which can be used as an alternative to the learned loss. The
first is a diversity loss $\lexp^\text{div}$ which encourages the agent to take varied actions. If the agent does happen to reach the same state multiple times it should definitely not repeat the action it previously took. Accordingly, 
\begin{equation} \label{eq:div}
    \lexp^\text{div}\left(\theta, \mathcal{D}_\tau^\text{int}\right) = 
    \sum_{i < j \leq k} g(s_{i},s_{j})\log \left(\pi_\theta^{\left(a_{i}\right)}(s_{j})\right),
\end{equation}
where $s_t$ is the agent's state at time $t$, $a_t$ is the action the agent takes
at time $t$, and $g$ calculates the similarity between two states. For simplicity we let $g(s_{i}, s_{j})$ be 1 if the pixel difference between $s_{i}$ and $s_{j}$ is below a certain threshold and 0 otherwise.

Additionally, we consider a prediction loss $\lexp^\text{pred}$ where the agent aims to predict the success of each action. The idea is to avoid taking actions that the network predicts will fail. We say that the agent's action has failed if we detect sufficient similarity in two consecutive states. This may occur when the agent bumps into an object or wall. In addition to producing a policy $\pi_\theta$ over actions the agent also predicts the success of each action. For state $s_t$ we denote the predicted probability that action $a$ succeeds as $q_\theta^{(a)}(s_t)$. Instead of sampling an action from $\pi_\theta(s_t)$ we instead use $\tilde \pi_\theta(s_t) = \pi_\theta(s_t) * q_\theta(s_t)$ where $*$ denotes element-wise multiplication. 

For $\lexp^\text{pred}$ we use a standard binary cross entropy loss between our success prediction $q_\theta^{(a)}$ and observed success. Using the same $g$ from Equation \eqref{eq:div} we write our loss as
\begin{equation} \label{eq:pred}
    \begin{split}
    \lexp^\text{pred}\left(\theta, \mathcal{D}_\tau^\text{int}\right) = 
    \sum_{t = 0}^{k-1} \mathcal{H}\left(q_{\theta}^{(a_t)}(s_t), 1-g(s_t,s_{t+1})\right),
    \end{split}
\end{equation}
where $\mathcal{H}(\cdot, \cdot)$ denotes binary cross-entropy. 

We acknowledge that in a non-synthetic environment it may be difficult to produce a reliable function $g$. Therefore we only use $g$ in the hand-crafted variations of the loss.

\begin{algorithm}[tp]
\caption{SAVN-Testing$(\mathcal{T}_\text{test}, \theta, \phi, \alpha, \beta, k)$}\label{alg:test}
\begin{algorithmic}[1]
\For{mini-batch of tasks $\tau_i \in \mathcal{T}_{\text{test}}$}
    \State{$\theta_i \gets \theta$}
    \State{$t \gets 0$}
    \While{termination action is not issued}
        \State{Take action $a$ sampled from $\pi_{\theta_i}(s_t)$}
        %\State{Run for $k$ steps to get self supervised exploration loss $\lexp$.}
        \State{$t \gets t + 1$}
        \If{$t$ is divisible by $k$}
        \State{$\theta_i \gets \theta_i - \alpha \nabla_{\theta_i}
                \lexp^{\phi}\left(\theta_i, \mathcal{D}_\tau^{(t,k)}\right)$}
        \EndIf
    \EndWhile
\EndFor
\end{algorithmic}
\end{algorithm}

\subsection{Training and Testing} \label{traininfer}

So far we have implicitly decomposed the agent's trajectory into an interaction and navigation phase. In practice, we would like the agent to keep adapting until the object is found during both training and testing. We therefore perform an SGD update with respect to the self-supervised interaction loss every $k$ steps. We compute the interaction loss at time $t$ by using the information from the previous $k$ steps of the agent's trajectory, which we denote $\mathcal{D}_\tau^{(t,k)}$. Note that $\mathcal{D}_\tau^{(t,k)}$ is analogous to $\mathcal{D}_\tau^{\text{int}}$ in Equation \eqref{eq:train}.  In addition, the agent should be able to navigate efficiently. Hence, we compute the navigation loss $\lnav$ using the the information from the complete trajectory of the agent, denoted by $\mathcal{D}_\tau$.

For the remainder of this work we refer to the gradient with respect to $\lexp$ as the
\emph{interaction-gradient} and the gradient with respect to  $\lnav$ as the \emph{navigation-gradient}. These gradients are illustrated in Figure \ref{fig:model} by red and green arrows, respectively. Note that we do not update the loss parameters $\phi$ via the interaction-gradient.

\begin{figure*}[tp]
    \centering
    \includegraphics[width=40pc]{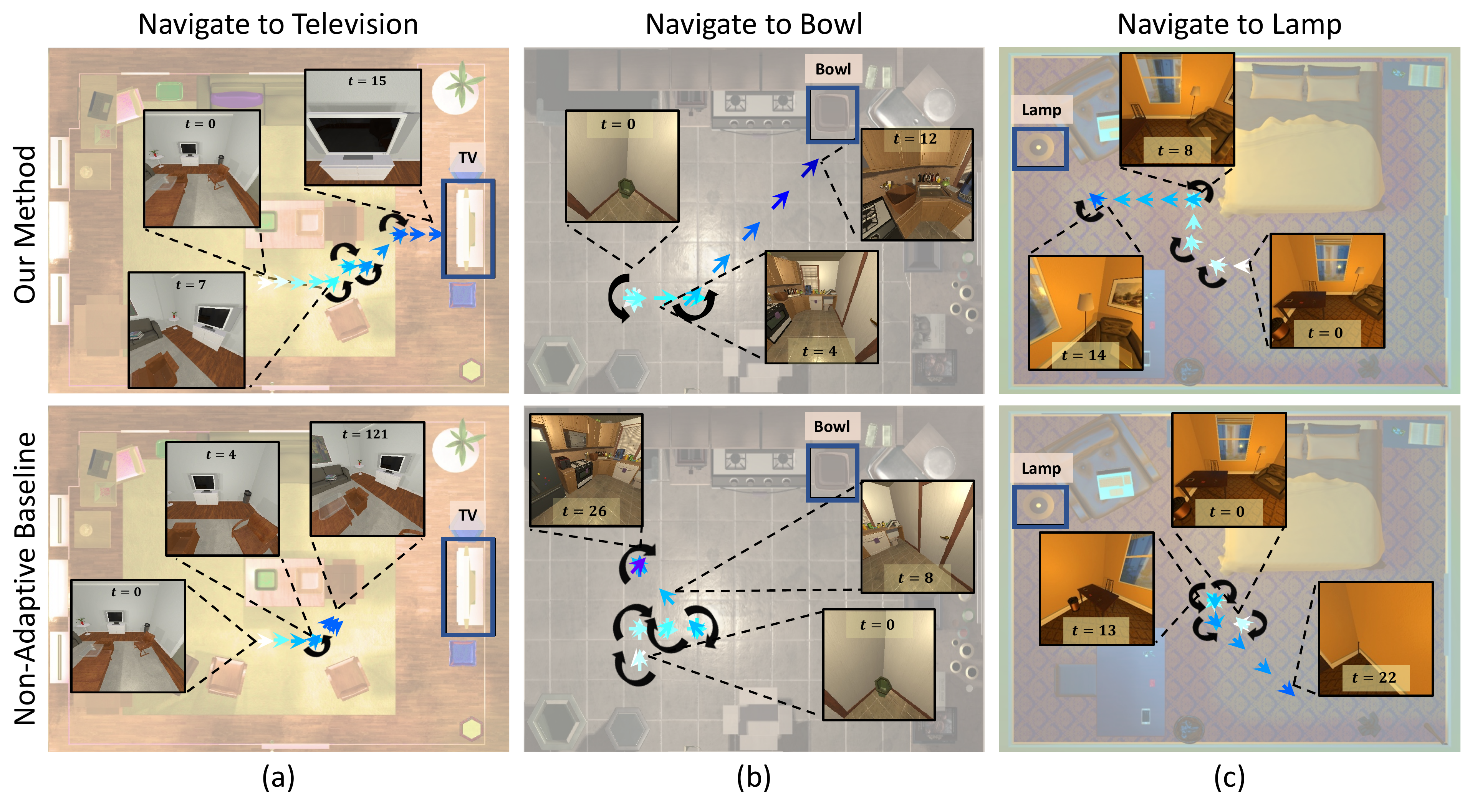}
    \vspace{-0.2cm}
    \caption{\textbf{Qualitative examples.} We compare our method with the non-adaptive baseline. We illustrate the trajectory of the agent (white corresponds to the beginning of the trajectory and dark blue shows the end). Black arrows represent rotation. We also show the egocentric view of the agent at a few time steps. Our method may learn from its mistakes (e.g., getting stuck behind an object).}
    \label{fig:qualitative}
\end{figure*}
Though traditional works use testing and inference interchangeably we may regard inference more abstractly as any setting in which the task is performed without supervision. This occurs not only during testing but also within each episode of navigation during training.

Algorithms \ref{alg:train} and \ref{alg:test} detail our method for training and testing, respectively. In Algorithm \ref{alg:train} we learn a policy network $\pi_\theta$ and a loss network parameterized by $\phi$ with step-size hyper-parameters $\alpha, \beta_1,\beta_2$. Recall that $k$ is a hyper-parameter which prescribes the frequency of the interaction-gradients. If we are instead considering a hand-crafted self-supervised loss then we ignore $\phi$ and omit line 12.

Recall that the adapted parameters, which we denote $\theta_i$ in Algorithm \ref{alg:train} and \ref{alg:test}, are implicitly a function of $\theta, \phi$. Therefore, the differentiation in lines 11 and 12 is well defined though it requires the computation of Hessian vector-products. We never compute more than $4$ interaction-gradients due to computational constraints.

At test time we may adapt in an environment with respect to the self-supervised interaction loss, but we no longer have access to $\lnav$. Note that the shared parameter $\theta$ is not updated during testing, as detailed in Algorithm \ref{alg:test}.

\section{Experiments}

Our goal in this section is to (1) evaluate our self-adaptive navigation model in comparison to non-adaptive baselines, (2) determine if the learned self-supervised objective provides any improvement over hand-crafted self-supervised losses, and (3) gain insight into how and why our method may be improving performance.

\subsection{Experiment setup}

We train and evaluate our models using the AI2-THOR \cite{ai2thor} environment. AI2-THOR provides indoor 3D synthetic scenes in four room categories, kitchen, living room, bedroom and bathroom. For each room type, we use 20 scenes for training, 5 for validation and 5 for testing (a total of 120 scenes).

We choose a subset of target object classes as our navigation targets such that (1) they are not hidden in cabinets, fridges, etc., (2) they are not too large that they take a big portion of the room and are visible from most parts of the room (e.g., beds in bedrooms). We choose the following sets of objects for each type of room: 1) Living room: pillow, laptop, television, garbage can, box, and bowl. 
2) Kitchen: toaster, microwave, fridge, coffee maker, garbage can, box, and bowl.
3) Bedroom: plant, lamp, book, and alarm clock.
4) Bathroom: sink, toilet paper, soap bottle, and light switch.

We consider the actions $\mathcal{A} =$ \{\texttt{MoveAhead}, \texttt{RotateLeft}, \texttt{RotateRight}, \texttt{LookDown}, \texttt{LookUp}, \texttt{Done}\}.  Horizontal rotation occurs in increments of 45 degrees while looking up and down change the camera tilt angle by 30 degrees. \done corresponds to the termination action discussed in Section \ref{sec:task}. The agent successfully completes a navigation task if this action is issued when an instance from the target object class is within 1 meter from the agent's camera and within the field of view. This follows from the primary recommendation of \cite{anderson18}. Note that if the agent ever issues the \done action when it has not reached a target object then we consider the task a failure.

\subsection{Implementation details}

We train our method and baselines until the success rate saturates on the validation set. We train one model across all scene types with an equal number of episodes per type using 12 asynchronous workers. For $\lnav$, we use a reward of 5 for finding the object and -0.01 for taking a step. For each scene we randomly sample an object from the scene as a target along with a random initial position. For our interaction-gradient updates we use SGD and for our navigation-gradients we use Adam \cite{kingma2014}. For step size hyper-parameters ($\alpha, \beta_1, \beta_2$ in Algorithm \ref{alg:train}) we use $10^{-4}$ and for $k$ we use 6. Recall that $k$ is the hyper-parameter which prescribes the frequency of interaction-gradients. We experimented with a schedule for $k$ but saw no significant improvement in performance. 

For evaluation we perform inference for 1000 different episodes (250 for each scene type). The scene, initial state of the agent and the target object are randomly chosen. All models are evaluated using the same set. For each training run we select the model that performs best on the validation set in terms of success.

\subsection{Evaluation metrics}

We evaluate our method on unseen scenes using both Success Rate and Success weighted by Path Length (SPL). SPL was recently proposed by \cite{anderson18} and captures information about navigation efficiency. Success is defined as 
$
    \frac{1}{N}\sum_{i=1}^N \mathcal{S}_i
$
and SPL is defined as
$
    \frac{1}{N}\sum_{i=1}^N \mathcal{S}_i \frac{L_i}{\max(P_i, L_i)}
$, 
where $N$ is the number of episodes, $\mathcal{S}_i$ is a binary indicator of success in episode $i$, $P_i$ denotes path length and $L_i$ is the length of the optimal trajectory to any instance of the target object class in that scene. We evaluate the performance of our model both on all trajectories and trajectories where the optimal path length is at least 5. We denote this by $L \geq 5$ ($L$ refers to optimal trajectory length).

\subsection{Baselines}

We compare our models with the following baselines: 

\noindent {\bf Random agent baseline.} At each time step the agent randomly samples an action using a uniform distribution.

\noindent {\bf Nearest neighbor (NN) baseline.} At each time step we select the most similar visual observation (in terms of Euclidean distance between ResNet features) among scenes in training set which contain an object of the class we are searching for. We then take the action that is optimal in the train scene when navigating to the same object class.

\noindent {\bf No adaptation (A3C) baseline.} The architecture for the baseline is the same as ours, however there is no interaction-gradient and therefore no interaction loss. The training objective for this baseline is then
$
    \min_{\theta} \sum_{\tau \in \mathcal{T}_\text{train}} \lnav\left(\theta, \mathcal{D}_\tau\right)
$
which is equivalent to setting $\alpha = 0$ in Equation \eqref{eq:train}. This baseline is trained using A3C \cite{mnih2016}.

\subsection{Results} \label{sec:results}
\begin{table}[tp]
	\centering
	\resizebox{\columnwidth}{!}{%
	\setlength{\tabcolsep}{1pt}
	\begin{tabular}{l|l l|l l|}
	    \cline{2-5}
	    
	   & \multicolumn{2}{|c|}{All} & \multicolumn{2}{|c|}{\lgf} \\ 
	   \cline{2-5}
	   & SPL & Success & SPL & Success \\
	    \hline
	    \multicolumn{1}{|c|}{Random} & $3.64_{(0.6)}$ & $8.0_{(1.3)}$ & $0.1_{(0.1)}$&$0.28_{(0.1)}$\\
	    \multicolumn{1}{|c|}{NN} &6.09& 7.90 &	1.38 &1.66 \\
	    \multicolumn{1}{|c|}{No Adapt (A3C)}& $14.68_{(1.8)}$ &$33.04_{(3.5)}$&$11.69_{(1.9)}$&$21.44_{(3.0)}$ \\
	    %\multicolumn{1}{|c|}{No Adaptation w/ mem}& 15.54& 34.30&		11.21&20.50\\
	    \multicolumn{1}{|c|}{Scene Priors \cite{yang18}} &$15.47_{(1.1)}$& $35.13_{(1.3)}$& $11.37_{(1.6)}$& $22.25_{(2.7)}$ \\
	    \multicolumn{1}{|c|}{Ours - prediction} &$14.36_{(1.1)}$&$38.06_{(2.9)}$&$12.61_{(1.3)}$&$26.41_{(2.4)}$\\
	    \multicolumn{1}{|c|}{Ours - diversity}& $15.12_{(1.5)}$& $39.52_{(3.0)}$& $13.38_{(1.4)}$& $27.66_{(3.5)}$\\
	    \multicolumn{1}{|c|}{Ours - \SAVN} &{\bf $\mathbf{16.15_{(0.5)}}$}&{\bf $\mathbf{40.86_{(1.2)}}$} &	{\bf $\mathbf{13.91_{(0.5)}}$}&{\bf $\mathbf{28.70_{(1.5)}}$}\\
	    \hline
	\end{tabular}
	}
	\caption{\textbf{Quantitative results.} We compare variations of our method with random, nearest neighbor and non-adaptive baselines. We consider two evaluation metrics, Success Rate and SPL. We provide results for all targets `All' and a subset of targets whose optimal trajectory length is greater than 5. We report the average over 5 training runs with standard deviations shown in sub-scripted parentheses.}
	\vspace{-0.3cm}
\label{tab:results}
\end{table}
Table~\ref{tab:results} summarizes the results of our approach and the baselines. We consider three variations of our method, which include SAVN (learned self-supervised loss) and the hand-crafted prediction and diversity loss alternatives. 

Our learned self-supervised loss outperforms all baselines by a large margin in terms of both success rate and SPL metrics. Most notably, we observe about 8\% absolute improvement in success and 1.5 in SPL over the non-adaptive (A3C) baseline. The self-supervised objective not only learns to navigate more effectively but it also learns to navigate  efficiently.

The models trained with hand-crafted exploration losses outperform our baselines by large margins in success, however, the SPL performance is not as impressive as with the learned loss. We hypothesize that minimizing these hand-crafted exploration losses are not as conducive to efficient navigation. 

\noindent {\bf Failed actions.} We now explore a behavior which sets us apart from the non-adaptive baseline. In the beginning of an episode the agent looks around or explores the free space in front of it. However, as the episode progresses, the non-adaptive agent might issue the termination action or get stuck. Our method (SAVN), however, exhibits this pattern less frequently.

\begin{figure}
    \centering
    \includegraphics[width=15pc]{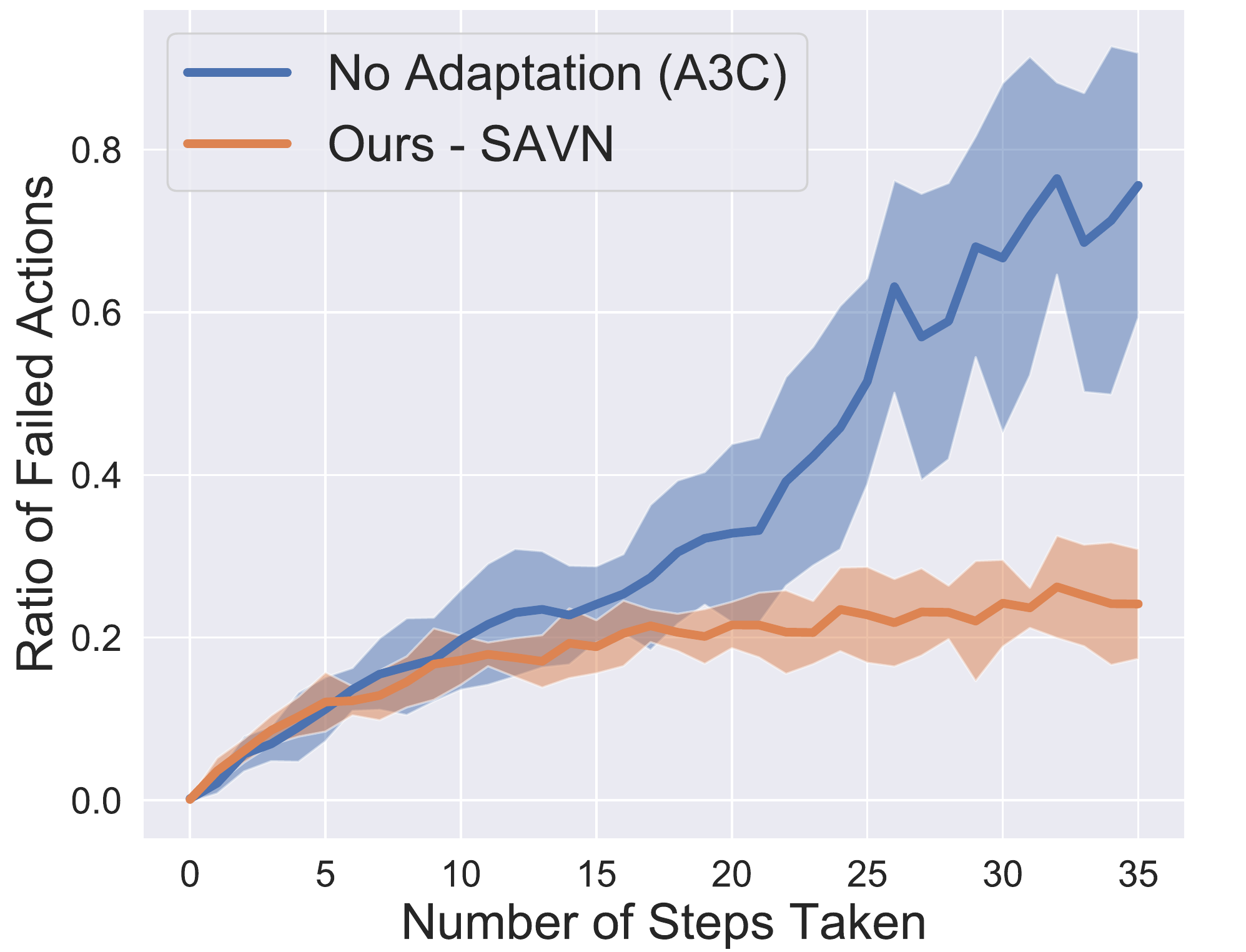}
    \caption{\textbf{Failed actions.} Our approach learns to adapt and not to take unsuccessful actions as the navigation proceeds.}
    \label{fig:failed}
    \vspace{-0.4cm}
\end{figure}

To examine this behavior we compute the ratio of actions which fail. Recall that an agent's action has failed if two consecutive frames are sufficiently similar. Typically, this will occur if an agent collides with an object. As shown in Figure \ref{fig:failed}, our method experiences significantly fewer failed actions than the baseline as the episode progresses.

\noindent {\bf Qualitative examples.} Figure~\ref{fig:qualitative} qualitatively compares our method with the non-adaptive (A3C) baseline. In scenario (a) our baseline gets stuck behind the box and tries to move forward multiple times, while our method adapts dynamically and finds the way towards the television. Similarly in scenario (c), the baseline tries to move towards the lamp but after bumping into the bed 5 times and rotating 9 times, it issues \done in a distant location from the target. 

\subsection{Ablation Study}

In this section we perform an ablation on our methods to gain further insight into our result.

\noindent {\bf Adding modules to the non-adaptive baseline.} In Table \ref{tab:vanillaablation} we experiment with the addition of various modules to our non-adaptive baseline. We begin by augmenting the baseline with an additional memory module which performs self-attention on the latest $k=6$ hidden states of the LSTM. SAVN outperforms the memory-augmented baseline as well.

Additionally, we add the prediction loss detailed in Section \ref{sec:learnedloss} to the training objective.
This experiment reveals that our result is not simply a consequence of additional losses. By using our training objective with the added hand-crafted prediction loss (referred to as `Ours - prediction'), we outperform the baseline non-adaptive model with prediction (referred to as `A3C w/ prediction') by 3.3\% for all trajectories and 4.8\% for trajectories of at least length 5 in terms of success rate. As discussed in the Section \ref{sec:results}, minimizing the hand-crafted objectives during the episode may not be optimal for efficient exploration. This may be why we show a boost in SPL for trajectories of at least length 5 but not overall. We run the same experiment with the diversity loss but find that the baseline model is unable to converge with this additional loss.

\begin{table}[tp]
	\centering
	\resizebox{\columnwidth}{!}{%
	\setlength{\tabcolsep}{2pt}
	\begin{tabular}{c|c c|c c|}
	    \cline{2-5}
	    
	   & \multicolumn{2}{|c|}{All} & \multicolumn{2}{|c|}{\lgf} \\ 
	   \cline{2-5}
	   & SPL & Success & SPL & Success \\
	    \hline
	    \multicolumn{1}{|c|}{No Adapt (A3C)}& $14.68$ &$33.04$&$11.69$&$21.44$ \\
	    \multicolumn{1}{|c|}{A3C w/ mem}& 15.54& 34.30&		11.21&20.50\\
	    \multicolumn{1}{|c|}{A3C w/ prediction loss}& 14.95&34.80&10.94&	21.60\\
	    \multicolumn{1}{|c|}{Ours - prediction} &$14.36$&$38.06$&$12.61$&$26.41$\\
	    \multicolumn{1}{|c|}{Ours - \SAVN} &{\bf $\mathbf{16.15}$}&{\bf $\mathbf{40.86}$} &	{\bf $\mathbf{13.91}$}&{\bf $\mathbf{28.70}$}\\
	    \hline
	    \hline
	    \multicolumn{1}{|c|}{A3C (GT obj)}&\hspace{0.14cm}31.34&\hspace{0.14cm}44.40&	\hspace{0.14cm}16.77&\hspace{0.14cm}26.05\\
	    \multicolumn{1}{|c|}{Ours - \SAVN (GT obj)} &\hspace{0.14cm}{\bf 35.55}&\hspace{0.14cm}{\bf 54.40}&	\hspace{0.14cm}{\bf 23.47}&\hspace{0.14cm}{\bf 37.87}\\
	    \hline
	\end{tabular}
	}
	\caption{\textbf{Ablation results.} We compare our approach with the non-adaptive baseline augmented with memory and our hand-crafted loss. We also provide the result when we use ground truth object information (bottom two rows).}
	\label{tab:vanillaablation}
\end{table}

\begin{figure}
    \centering
    \includegraphics[width=20pc]{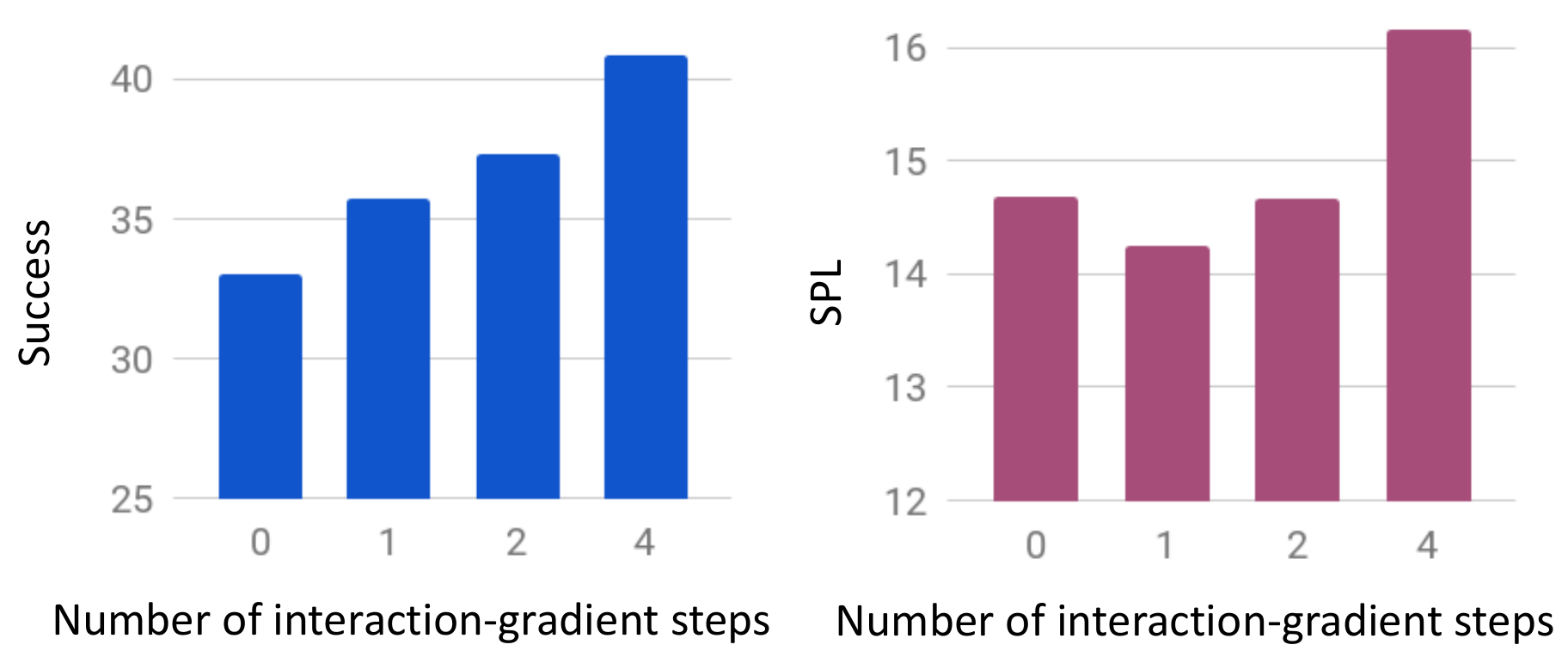}
    \caption{\textbf{Number of Gradients Ablation.} Our success rate increases as more interaction-gradients are taken during training/testing.}
    \label{fig:gl}
    \vspace{-0.4cm}
\end{figure}

\noindent {\bf Ablation of the number of gradients.} To showcase the efficacy of our method we modify the number of interaction-gradient steps that we perform during the adaptation phase during training and testing. As discussed in Section \ref{traininfer}, we never perform more than 4 interaction-gradients due to computational constraints. As illustrated by Figure \ref{fig:gl}, there is an increase in success rate when more gradient updates are used,  demonstrating the importance of the interaction-gradients.

\noindent {\bf Perfect object information.} Issuing the termination action at the correct location plays an important role in our navigation task. We observe that \SAVN still outperforms the baseline even when the termination signal is provided by the environment (referred to as `GT obj' in Table~\ref{tab:vanillaablation}).

\section{Conclusions}

We introduce a self-adaptive visual navigation agent (SAVN) that learns during both training and inference. During training the model learns a self-supervised interaction loss that can be used when there is no supervision. Our experiments show that this approach outperforms non-adaptive baselines by a large margin. Furthermore, we show that the learned interaction loss performs better than hand-crafted losses. Additionally, we find that \SAVN navigates more effectively than a memory-augmented non-adaptive baseline. We conjecture that this idea may be applied in other domains where the agents may learn from self-supervised interactions. 

\small{
\noindent\textbf{Acknowledgements:} We thank Marc Millstone and the Beaker team for providing a robust experiment platform and providing tremendous support. We also thank Luca Weihs and Eric Kolve for their help with setting up the framework, Winson Han for his help with figures, and Chelsea Finn for her valuable suggestions. This work is in part supported by NSF IIS-165205,  NSF IIS-1637479, NSF IIS-1703166, Sloan Fellowship, NVIDIA Artificial Intelligence Lab, and Allen Institute for artificial intelligence.  
}

{\small
\bibliographystyle{ieee_fullname}
\bibliography{egbib}
}

\end{document}